\documentclass[letterpaper, 10 pt, conference]{ieeeconf}
\IEEEoverridecommandlockouts                              
\usepackage{times}
\usepackage{newtxtext,newtxmath}
\usepackage{breqn}
\usepackage{amsfonts}
\usepackage{amsmath}
\usepackage{comment}
\usepackage{graphicx}
 \usepackage{hyperref} 
 \usepackage{algorithm}
\usepackage{algpseudocode}
\usepackage{booktabs}  
\usepackage{array}     
\usepackage{multirow}
\usepackage{caption}
\usepackage{subcaption}
\usepackage{afterpage}

\newcommand*\dif{\mathop{}\!\mathrm{d}}

\newcommand{\norm}[1]{\left\lVert#1\right\rVert}

\begin{document}

\date{}

\title{\Large\bf Odometry Without Correspondence from Inertially Constrained Ruled Surfaces}



\author{Chenqi Zhu$^{1}$, Levi Burner$^{2*}$, Yiannis Aloimonos$^{2,3}$
\thanks{$^{1}$ Research at Panasonic Connect. This work was completed while author was a student in Computer Science at the University of Maryland, College Park and is not affiliated with Panasonic.
        {\tt\small zhu.chenqi@jp.panasonic.com}}
\thanks{$^{2*}$ Corresponding Author. Department of Computer Science, University of Maryland, College Park,
        {\tt\small lburner@umd.edu}}
\thanks{$^{3}$ University of Maryland Institute for Advanced Computer Studies, University of Maryland, College Park
        {\tt\small jyaloimo@umiacs.edu }}
}

\maketitle

\section*{\centering Abstract}

{\em
Visual odometry techniques typically rely on feature extraction from a sequence of images and subsequent computation of optical flow. This point-to-point correspondence between two consecutive frames can be costly to compute and suffers from varying accuracy, which affects the odometry estimate's quality.
Attempts have been made to bypass the difficulties originating from the correspondence problem by adopting line features and fusing other sensors (event camera, IMU) to improve performance, many of which still heavily rely on correspondence.
If the camera observes a straight line as it moves, the image of the line sweeps a smooth surface in image-space time. It is a ruled surface and analyzing its shape gives information about odometry. Further, its estimation requires only differentially computed updates from point-to-line associations. Inspired by event cameras' propensity for edge detection, this research presents a novel algorithm to reconstruct 3D scenes and visual odometry from these ruled surfaces.
By constraining the surfaces with the inertia measurements from an onboard IMU sensor, the dimensionality of the solution space is greatly reduced.
}

 \begin{figure*}[!htb]
    \centering
    \begin{subfigure}[t]{0.45\textwidth}
        \centering
        \includegraphics[width=\textwidth]{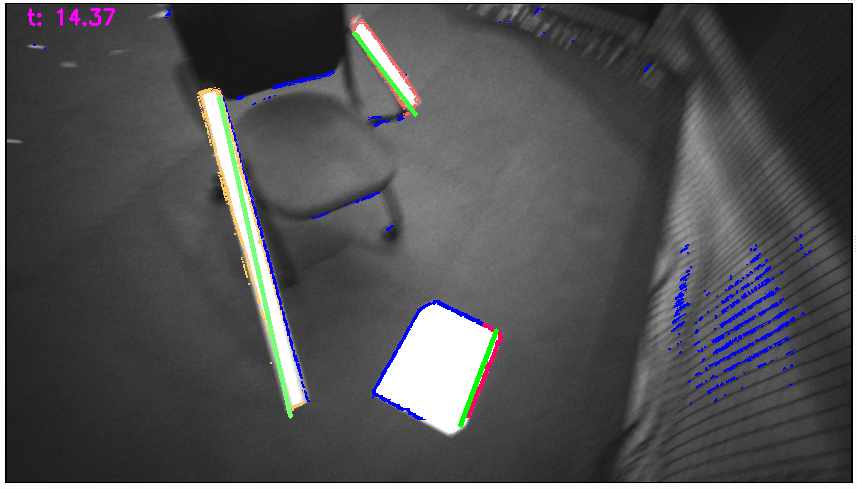}
        \caption{Rulings projected to image frame. The blue points are thresholded image gradient pixels, and the green lines represent rulings projected to image frame. Pixels within a certain distance of the rulings are colored in red and yellow and are used as data points for ruled surfaces estimation.}
        \label{fig:1}
    \end{subfigure}
    \hfill
    \begin{subfigure}[t]{0.45\textwidth}
        \centering
        \includegraphics[width=\textwidth]{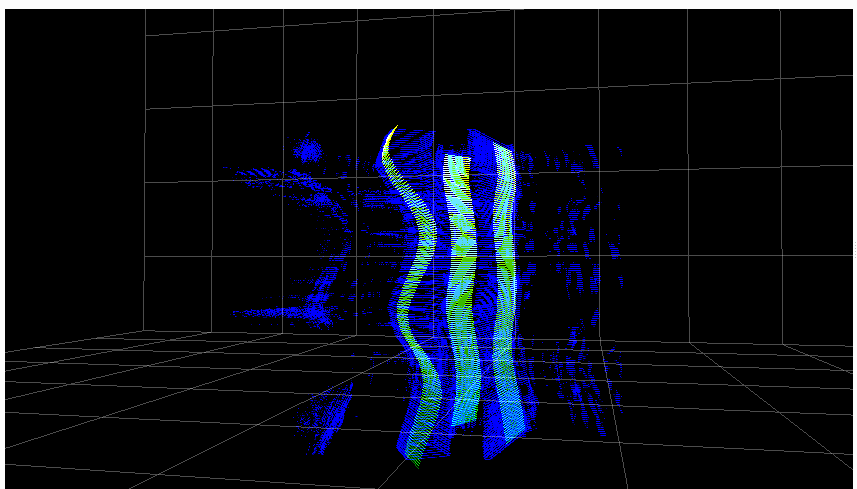}
        \caption{Surfaces $TS_{X_0, V_0} (t, \alpha)$ ruled by projections of lines in image-time space, approximated in a $k=140$-length window (roughly 1.5 seconds). The $z$-axis (vertical up) is time $t$.}
        \label{fig:2}
    \end{subfigure}
    \caption{(a) rulings projected to image frame and (b) ruled surfaces in image-time space.}
    \label{fig:1_2_combined}
\end{figure*}

\section{Introduction}
The task of odometry estimation is of great significance.
An autonomous agent's abilities to localize, such as determining the position of drones or vehicles relative to buildings or street signs, have consequential real world applications.
This set of problems, where a 3D structure of the environment along with the motion signal of the agent is reconstructed from observations of 2D image sequence, is commonly referred to as Structure from Motion (SfM).
It is traditionally solved with dense motion fields (optical flow) or triangulation via epipolar geometry. 
By identifying feature points between consecutive frames in  a sequence of images and calculate the displacement between each pair of corresponding points, optical flow can be constructed as an approximation of the motion field from which the 3D motion can be recovered.
While in triangulation via epipolar geometry, a 3D point in space is reconstructed by corresponding two image points taken from different views whose relationship is defined by an essential matrix.

All these strategies that rely heavily on the correctness of point correspondence have their shortcomings as their accuracies are often limited by the quality of feature extractors \cite{Fermuller2000R}. Many methods employ bundle adjustment \cite{Fitzgibbon1998}, which attempts to simultaneously estimate the relative motion and scene geometry.
This approach is most prominent in its application in Simultaneous localization and mapping (SLAM) and Visual Odometry, where both the mapping of an unknown environment and the tracking of the camera locations are recovered simultaneously \cite{Davison2003}. 
But this method still depends on the abundance of features for appropriate point correspondence, the lack of which commonly challenges the quality of the algorithm in the wild. One famous incident being NASA's Ingenuity Mars helicopter crashed due to  featureless terrain \cite{Evan2024}.

In recent years, neural networks have been developed to leverage the power of deep learning to solve camera pose \cite{Kendall2015}.  
Rather than seeking to match the features explicitly, neural networks are typically capable of approximating optical flow by learning the hidden representation through their enormous size of parametrization. To offer more explainable results, works have also be done to incorporate additional constraints stemming from the physical properties of the motion field, like normal flow, into their network \cite{Parameshwara2022}.
However, these results are often computationally costly, lack theoretical guarantees, and are susceptible to over-fitting, making generalization difficult. Finally, due to computational cost, they can only consider visual information over small intervals such as 10's of milliseconds, which greatly increases the difficulty of the odometry estimation problem.
Here the presented research argues that the problem of visual odometry can be solved through a low dimensional problem that differentially updates instead of discretely corresponding to independently detected features. Its dimensionality scales in the number lines but is fixed in the length of time considered by the sliding window solver.

This work is motivated by the fact that our world is filled with straight lines, whether it be the edges of tables or outlines of buildings.
Previous research that worked on structure from motion with lines more or less viewed them as easy features for correspondence \cite{Baker2004, Baker2005, Bartoli2005, Schindler2006, Shin2024}. 
While the line geometry yields adequate assistance, this work shows that the continuous motion of lines in image-space time has a special geometric property, they form a type of surface, called a ruled surface, which describes the 3D scene and can be used to derive the odometry of the observing agent.
This work introduces an inertially constrained model based on ruled surfaces that enables the computation of visual odometry and reconstruction of 3D scenes with low-dimensional solutions, by taking advantage of the geometric properties of the ruled surface to avoid point-to-point or line-to-line correspondence.
A reprojection loss with an inner optimization solved in closed form is developed which estimates the optimal parameters subject to those geometric constraints. 
At all stages, the possible ruled surfaces are constrained by the IMU, thus realizing an estimator with tight visual-inertial coupling despite the relaxation of point correspondence, and are estimated in a sliding window fashion. 

\begin{figure*}[!htb]
    \centering
    \begin{subfigure}[t]{0.45\textwidth}
        \centering
        \includegraphics[width=\textwidth]{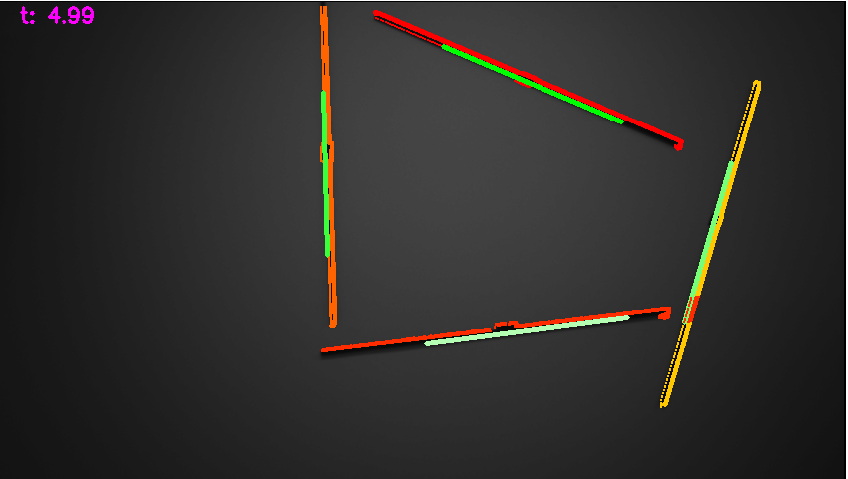}
        \caption{Scene used for experiments in Section \ref{sec:r1}. Four coplanar lines are present in the scene. Estimated lines are in green, and the sampled image points belonging to the ruled surfaces of each line are shown in warm colors.}
        \label{fig:10_31_028_1}
    \end{subfigure}
    \hfill
    \begin{subfigure}[t]{0.45\textwidth}
        \centering
        \includegraphics[width=\textwidth]{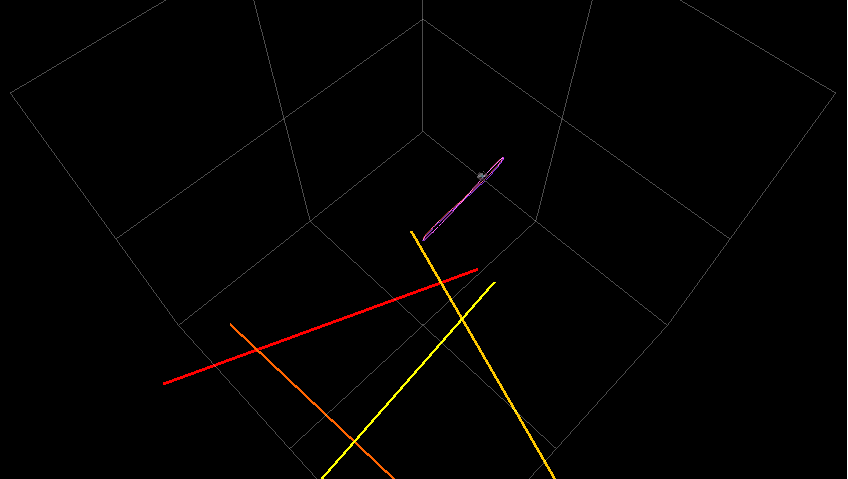}
        \caption{Estimated trajectory of the camera and estimated poses of lines in the scene in world coordinates. Lines are in warm colors, and the trajectory is shown in purple, indicating linear motion parallel to the scene plane.}
        \label{fig:10_31_028_4}
    \end{subfigure}
    
    \caption{(a) Scene setup with coplanar lines and estimated rulings, and (b) the camera trajectory alongside line poses in world coordinates.}
    \label{fig:10_31_028_1_10_31_028_4}
\end{figure*}

\begin{figure}[!htb]
    \centering
    \includegraphics[width=0.5\textwidth]{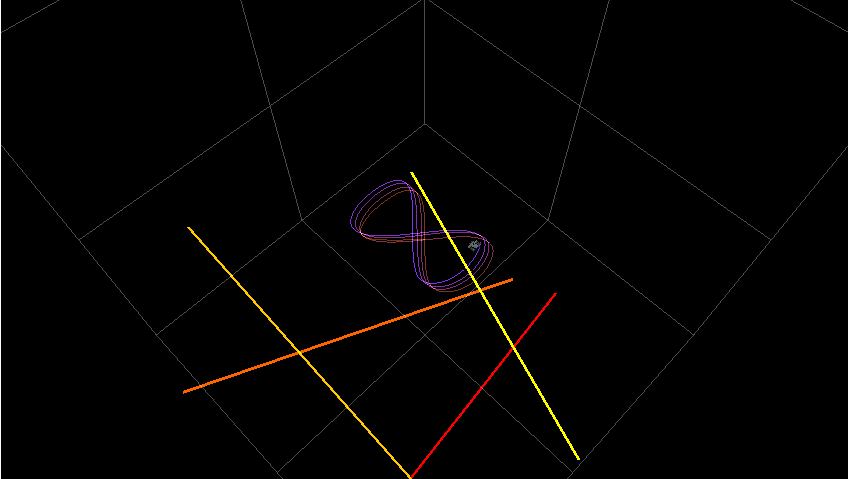}
    \caption{Estimated trajectory of the camera and the
estimated poses of the lines in the scene in world coordinates.
Lines are in warm colors and the trajectory in purple. Circular
motion parallel to the scene plane.}
    \label{fig:11_04_008_4}
\end{figure}

\begin{figure}[!htb]
    \centering
    \includegraphics[width=0.5\textwidth]{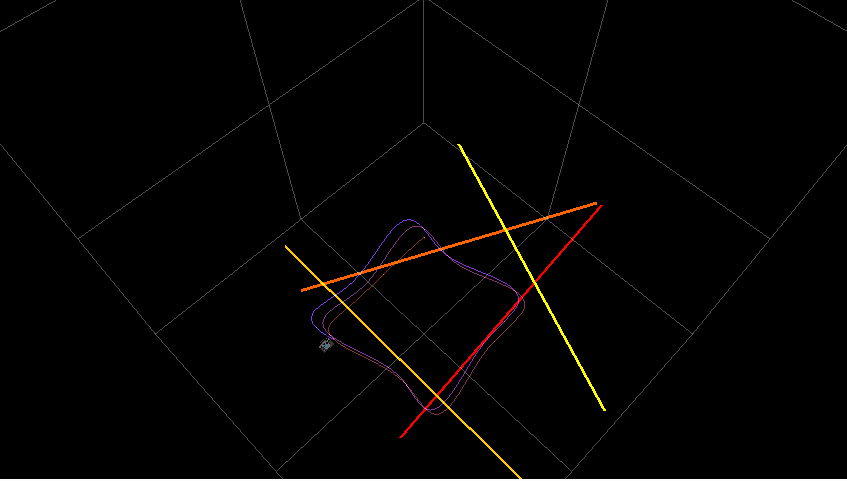}
    \caption{Estimated trajectory of the camera and the
estimated poses of the lines in the scene in world coordinates.
Lines are in warm colors and the trajectory in purple. Square non-smooth
motion parallel to the scene plane.}
    \label{fig:11_05_006_4}
\end{figure}

\section{Previous Work}

The classical way to solve structure from motion is to consider the problem as calculating the relative 3D motion from precise measurements on 2D images. 
With the spatio-temporal derivatives of two consecutive frames, either point correspondence \cite{Heyden2005} or optical flow \cite{Fermuller1997}, are first computed and the epipolar constraints are applied to gauge the 3D motion. 
This point correspondence is indispensable to these approaches, and the errors in point correspondence as a result of failure of feature detectors can exacerbate the errors in 3D motion estimation. 
While it is possible to improve the quality of point matching and optical flow, it is also possible to introduce methods to provide additional guarantees on the subsequentl estimated 3D motion.  
Works have exploited the fact that the projection of optical flow in the direction of image gradient (normal flow) is resilient to outliers to enforce robust and direct constraints on the estimated camera pose \cite{Fermuller1995, Parameshwara2022}. 
Moreover, other works have proposed practices to discard the erroneous optical flow and recover motion from normal flow directly \cite{Barranco2021, Ren2024}.
Additional constraints are also available when considering the motion field. The epipolar constraint can be introduced by measuring the deviation in epipolar view and the positive depth constraint, which is derived from the fact that the scene has to be in front of the image to be visible, can be applied to normal flow \cite{Fermuller1998, Fermuller2000}. 

While the normal flow based techniques theoretically overcome the challenges of pure optical flow based ones, more can be done to better estimate odometry. 
When considering recovering motion from visual inputs, the geometric properties of the observed scene over long time intervals offer valuable information for solving the problem. One such example is the surfaces carved by straight lines in the world projected into image space.
Line-correspondence has been proposed as an alternative to point-correspondence with a closed form solution for camera motion \cite{Spetsakis1990} Lines are not as straightforward to use as points due to their underconstrained nature.
Nevertheless, with the use of Pl\"ucker's coordinates, spatial relationships of lines can be derived relatively easily \cite{Plucker1865, Shevlin1998}.
Algorithms that solve structure from motion with lines calculate the camera pose under the epipolar constraint but with the properties of lines, either through interpreting parallel lines as textures \cite{Baker2004, Baker2005} or line-to-line correspondence \cite{Bartoli2005, Schindler2006}. 
Newer hardware like RGB-Depth camera also enabled camera extrinsic approximation from line features \cite{Shin2024}. 

Additionally, the  information that can be retrieved from other sensors are crucial to addressing visual odometry problems.
In the case of self-motion, IMU sensors can provide the angular velocity and acceleration up to a gravitational bias.
For instance, PL-VIO combines the properties of lines and inertial odometry, namely inertial measurements enabled by IMU sensor, to solve the visual odometry problem using point and line correspondences with a tight coupling to the IMU \cite{He2018}.
Their approaches unify the point features and line features while making use of the inertial data; a joint optimization is set up to minimize the IMU measurement residuals and re-projection residuals.

Recent advancements in visual sensors also help expand the types of visual inputs available. 
Event cameras are a special type of vision sensor that captures the change of brightness in images asynchronously. 
This unique property has led to more robust computation of optical flow by analyzing the time surface spanned by events \cite{Nagata2021}, which is then integrated into solving egomotion estimation, either through a combination of event camera and traditional camera \cite{Yang2022} or through deep learning \cite{Zhu2018}.
Further, due to the nature of the sensor, it is easier to identify and cluster events belonging to the same edge. 
Utilizing this characteristic of event camera and line geometry, \cite{Chamorro2022} proposed event-based line-SLAM that works in the fashion of parallel tracking and mapping.
By combining it with IMU measurements, the linear solver was used to estimate motion with event camera. For a small enough time interval and assuming constant linear velocity, events generated by a line circumscribe a manifold from which the motion can be recovered \cite{Gao2023, Gao2024}. 
\cite{Le_Gentil_2020} extracts line positions from temporal event clouds through minimizing event-to-line distance in the image projections. 

Our approach goes beyond treating lines in each images as individual features for correspondence or parameterizing their motion over short times. 
Instead, it observes that a straight line sweeps a special type of surface, ruled surface, as it moves. 
By inspecting the shapes of ruled surfaces, visual odometry can be directly reconstructed without any assumption on its motion.

\section{Background}
In geometry, a surface $S$ is ruled if for every point $p$ on the surface there's a straight line $l$ on $S$ that also passes through $p$. 
Formally, a ruled surface is a mapping $F_{(\gamma, \delta)} : I \times \mathbb{R} \rightarrow \mathbb{R}^3$ defined as 

 \begin{equation}\label{eq:0} F_{(\gamma, \delta)} (t, \alpha) = \boldsymbol{\gamma}(t) + \alpha \boldsymbol{\delta}(t) \end{equation}

Where directrix $\boldsymbol{\gamma}(t) : I \rightarrow \mathbb{R}^3$, ruling direction $\boldsymbol{\delta}(t) : I \rightarrow \mathbb{R}^3 \setminus \{\boldsymbol{0}\}$ are smooth mappings on open interval $I$. At each fixed time $t$, surface $F_{(\gamma, \delta)}$ is ruled by a straight line that only depends on $\alpha$: $\boldsymbol{u}(\alpha) = \boldsymbol{\gamma}(t) + \alpha \boldsymbol{\delta}(t)$  \cite{Izumiya2003}.

For some fixed $t_0$ at which the observation starts, let $X_0 = \boldsymbol{\gamma}(t_0)$ and $V_0 = \boldsymbol{\delta}(t_0)$, a line in $\mathbb{R}^3$ parameterized by only $\alpha$ is \begin{equation}
X(\alpha) = X_0 + \alpha V_0
\end{equation}

For a rotation-less camera (stabilized by an IMU sensor), if the camera translates relative to that 3D line according to signal $-X(t)$, for a rigid environment, the motion of the line relative to camera is $X(t)$. This subsequently leads to a line that translates in $4$-dimensional space-time space. The surface ruled by such line in camera frame is 

\begin{equation} \label{eq:3}
S_{X_0, V_0} (t, \alpha) = X_0 + \alpha V_0 + X(t)
\end{equation}

For objects in the camera field of view, the projection of such line at time $t$ and displacement $\alpha$ to 2D image point $p$ is 
\begin{equation}  \label{eq:4}
p(t, \alpha) = \frac{X_0 + \alpha V_0 + X(t)}{\left[ X_0 + \alpha V_0 + X(t)\right]^z}
\end{equation}

Since the projection of a $3D$ line is still a line in image plane (except when the line is perpendicular to the image plane), it naturally entails that the surface swept by the projected line in image-time space is also ruled. To differentiate it from \eqref{eq:3}, below is defined as ruled image-surface which is a surface in $3$-dimensional image-time space

\begin{equation} \label{eq:5}
TS_{X_0, V_0} (t, \alpha) = \begin{bmatrix}
    p(t,a)\\
    t
\end{bmatrix}
\end{equation}

It is evident that $TS_{X_0, V_0}$ can be constructed by $S_{X_0, V_0}$ but the converse remains to be solved. Hence the following formulation entails. 

\begin{figure*}[!htb]
    \centering
    \begin{subfigure}[t]{0.32\textwidth}
        \centering
        \includegraphics[width=\textwidth]{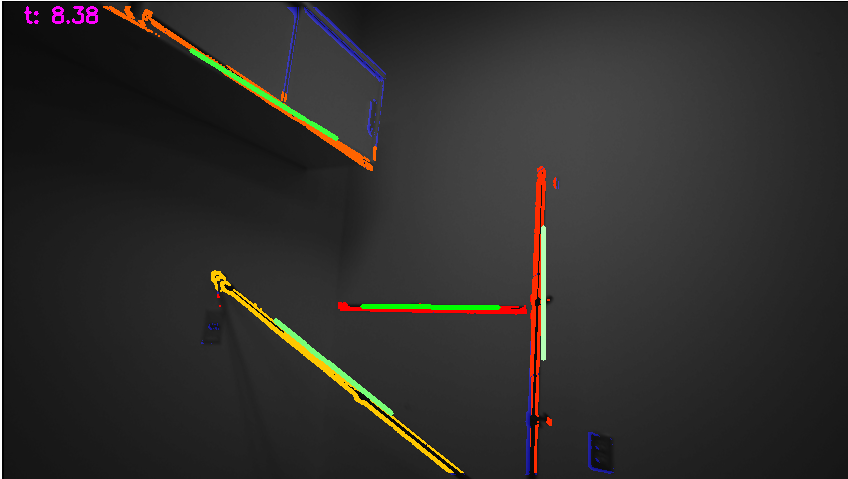}
        \caption{Scene used for experiments in Section \ref{sec:r4}. Four non-coplanar lines are tracked in this scene.}
        \label{fig:11_06_008_1}
    \end{subfigure}
    \hfill
    \begin{subfigure}[t]{0.32\textwidth}
        \centering
        \includegraphics[width=\textwidth]{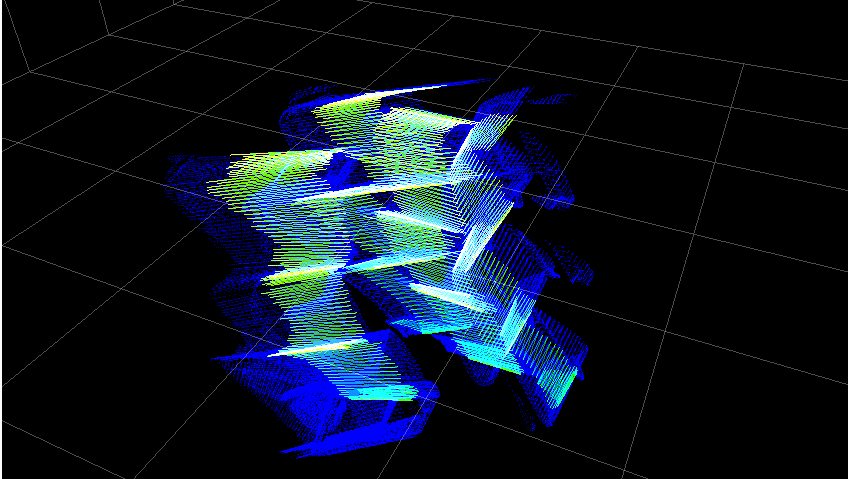}
        \caption{The non-coplanar scene generates a much more complicated point cloud in image-time space, making fitting surfaces difficult.}
        \label{fig:11_06_008_2}
    \end{subfigure}
    \hfill
    \begin{subfigure}[t]{0.32\textwidth}
        \centering
        \includegraphics[width=\textwidth]{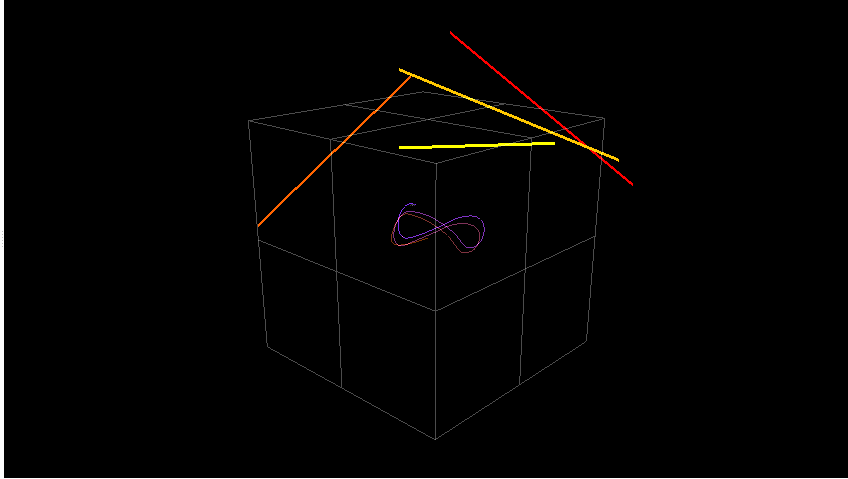}
        \caption{Estimated trajectory of the camera and the
estimated poses of the lines in the scene in world coordinates.
The latter are clearly not coplanar.}
        \label{fig:11_06_008_4}
    \end{subfigure}
    
    \caption{(a) Scene setup with non-coplanar lines, (b) point cloud and estimated ruled surface in image-time space, and (c) the camera trajectory alongside line poses.}
    \label{fig:11_06_008_all}
\end{figure*}

\section{Problem Formulation}  \label{sec:4}

Given a set of 2D data points from a sequence of images $\{ p_i \}_{i=1}^{N}$ (whose selection procedure is explained in Section \ref{method}) that presumably belong to a surface, $TS_{X_0, V_0} (t, \alpha)$, which is ruled by the projection of a line $l$ in $\mathbb{R}^3$ , the objective is to approximate $TS$, from which a surface $S$ ruled by the motion of $l$ in camera frame can be deduced. That will ultimately yield the camera translation signal $-X(t)$, which is the camera motion relative to the ruling $l$. 

Thus, the focus is on setting up an optimization problem with a parameterization of $TS_{X_0, V_0} (t, \alpha)$. Start by rearranging equation \eqref{eq:4}  
\begin{equation} \label{eq:6}
\left[ X_0 + \alpha V_0 + X(t)\right]^z p(t, \alpha) = X_0 + \alpha V_0 + X(t)
\end{equation}

If the linear acceleration measurements from the IMU sensor is available, $X(t)$ is a function of six parameters

\begin{equation}
X(t) = X(t_0) + \dot X (t_0) (t-t_0) + \int_{t_0}^{t} \int_{t_0}^{\sigma} \ddot X(\sigma_2) \dif\sigma_2 \dif\sigma_1
\end{equation}

An additional term of $\frac{1}{2}t^2 g$ may be needed if only an IMU with gravitational bias is available. In practice it is observed that an additional linear bias $\xi$ is introduced by the integration error, which can be lumped together with $\ddot X$ during optimization.

Define operator $\Gamma\{\ddot X\} = \int_{t_0}^{t} \int_{t_0}^{\sigma} \ddot X(\sigma_2) \dif\sigma_2 \dif\sigma_1$ and assume the sequence starts at $t_0 = 0$

\begin{equation} \label{eq:8}
X(t) = X(0) + (\dot X (0) + \xi)t + \Gamma\{\ddot X\} + \frac{1}{2}t^2 g
\end{equation}

Combining equation \eqref{eq:5} and equation \eqref{eq:8}, note the displacement $X(0)=0$

\begin{dmath}\label{eq:9}
    \left[X_0^z + \alpha V_0^z + t(\dot X^z(0)+\xi^z) + \Gamma\{\ddot X^z\} + \frac{1}{2}t^2 g^z \right] p(t, \alpha) \\
    = \left[X_0^{(x, y)} + \alpha V_0^{(x, y)} + t(\dot X^{(x, y)}(0)+\xi^{(x, y)}) + \\ \Gamma\{\ddot X^{(x, y)}\} + \frac{1}{2}t^2 g^{(x, y)} \right]
\end{dmath}

Lumping the terms by their common multipliers and substitute the symbols gives the parameterization by $a,b,c,d,e,f,g$, where $d,e,f$ are two dimensional while $a,b,c$ are scalars. 

\begin{dmath}\label{eq:10}
 p(t, \alpha) =
\resizebox{0.5\textwidth}{!}{$
    \frac{ \alpha \overbrace{V_0^{(x, y)}}^d + \overbrace{X_0^{(x, y)}}^e + t\overbrace{(\dot X^{(x, y)}(0) + \xi^{(x, y)})}^f + \Gamma\{\ddot X^{(x, y)}\} + \frac{1}{2}t^2 g^{(x, y)} }{\underbrace{X_0^z}_a+ t \underbrace{(\dot X^z(0) + \xi^z)}_b + \alpha \underbrace{V_0^z}_c + \Gamma\{\ddot X^{z}\} + \frac{1}{2}t^2 g^z }$}
\end{dmath}

Where the directrix at $t_0$ is $X_0$, the ruling direction vector is $V_0$, and the camera translation signal is $-X(t)$. They are the components of \eqref{eq:3} which can now be constructed from $TS_{X_0, V_0} (t, \alpha)$. 

\begin{equation} \label{eq:11}
X_0 = \begin{bmatrix}
e^x\\
e^y\\
a\\
\end{bmatrix}, 
V_0 = \begin{bmatrix}
d^x\\
d^y\\
c\\
\end{bmatrix},
X(t) = \begin{bmatrix}
tf^x + \Gamma\{\ddot X^{x}\} + \frac{1}{2}t^2 g^x\\
tf^y + \Gamma\{\ddot X^{y}\} + \frac{1}{2}t^2 g^y\\
tb + \Gamma\{\ddot X^{z}\} + \frac{1}{2}t^2 g^z\\
\end{bmatrix}
\end{equation}

This parameterization gives six unknowns for the ruling, three unknowns for the camera velocity plus bias, and three unknowns for gravitational bias. An unbounded line in 3D, however, can be reduced to only having 4 degrees of freedom, as it can translate and rotate along itself. 

Therefore,  additional constraints to bound the dimensionality of the solution space are needed. Forcing $V_0$ to be a unit vector and $X_0$ to be the point such that the vector from origin to $X_0$ is perpendicular to the line essentially removes two degrees of freedom and reflects the line’s four degrees of freedom in the 3D space. In addition, the observed points to be in front of the camera which makes the denominator in the right-hand side strictly greater than $0$. 

\subsection{Ruling Reprojection Loss}\label{sec:loss}

Let $\{ p_i \}_{i=1}^{N}$ be the set of points that are projected by some ruling with $X_0$ a point on an unbounded line in 3D with direction $V_0$, it is desired to find optimal parameters that minimizes this error of ruling reprojection, which is the difference between left-hand side and right-hand side of equation \eqref{eq:4}, or rather its parameterized equivalent equation \eqref{eq:10}, bounded by other constraints. 

Formally, define the ruling reprojection loss at time $t$ to be the sum of the squared difference between projected points and estimated ruling with parameters $\boldsymbol{\theta} = (a,b,c,d,e,f,g)$

\begin{dmath}\label{eq:12}
   \mathcal{L}_t(\boldsymbol{\theta}) = \sum\limits_{i=1}^{N} \left|\left| p_i - \frac{\alpha_i d + e + tf + \Gamma\{\ddot X^{(x, y)}\} + \frac{1}{2}t^2 g^{(x, y)}}{a + tb + \alpha_i c + \Gamma\{\ddot X^{z}\} + \frac{1}{2}t^2 g^z}
    \right| \right|^2
\end{dmath}

Another thing remained to be tackled is the implicit displacement $\alpha_i$ along the ruling for each point $p_i$. They are dependent on each other. In the real world, some unknown $\alpha_i$ evaluates a point $p_i$ through projection while in the optimization problem $p_i$ is known and we want to estimate an $a_i$. This is equivalent to solving 

\begin{dmath}\label{eq:13}
    \min_{\alpha_i} \left| \left| p_i \left(a + tb + \alpha_i c + \Gamma\{\ddot X^{z}\} + \frac{1}{2}t^2 g^z \right) -  \left( \alpha_i d + e + tf + \Gamma\{\ddot X^{(x, y)}\} + \frac{1}{2}t^2 g^{(x, y)} \right) \right|\right|^2
\end{dmath}

By switching variables between its left-hand-side and right-hand-side, it can be solved with a system of $2$ linear equations.

\begin{dmath} \label{eq:14}
\begin{matrix}
(p_i^xc-d^x)\alpha_i  = \Phi_i^x\\
(p_i^yc-d^y)\alpha_i = \Phi_i^y
\end{matrix}
\end{dmath}

Where \begin{dmath}
    \Phi_i^w = \left[ e^w + tf^w + \Gamma\{\ddot X^{w}\} + \frac{1}{2}t^2 g^{w} - p_i^w \left( a + tb + \Gamma\{\ddot X^{z}\} + \frac{1}{2}t^2 g^z \right) \right]
\end{dmath}

Now, the whole constrained minimization problem became

\begin{equation}\label{eq:15}
   \min_{\boldsymbol{\theta}} \mathcal{L}_t(\boldsymbol{\theta})
\end{equation}

$s.t. \qquad \norm{V_0} = 1$

$\qquad \qquad X_0 \cdot V_0 = 0$

$\qquad \qquad a + tb + \alpha_i c + \Gamma\{\ddot X^{z}\} + \frac{1}{2}t^2 g^z > 0, \forall i$

$where \quad\alpha_i = (P_i^{\top}P_i)^{-1}_iP_i^{\top}\Phi_i$

$\qquad \qquad P_i = \begin{bmatrix}
    p_i^xc-d_x\\
    p_i^yc-d_y
\end{bmatrix}, \Phi_i = \begin{bmatrix}
    \Phi_i^x\\
    \Phi_i^y
\end{bmatrix}$

\subsection{Dimensionality}\label{sec:4_2}

While the problem is solvable for one surface, the estimated pose would be unbounded along the direction of the ruling of that surface. Thus to restore the unique camera pose, a minimum number of $2$ non-parallel lines are necessary.

For $M$ lines ($M > 1$), assuming static environment, the parameters $b,f$, and gravitational bias $g$ can be shared across different lines. Each surface would have 6 independent parameters and 6 shared parameters given us $6M+6$ unknowns. As discussed, an unbounded line in 3D have only 4 degrees of freedom. Thus, under the additional constraints, the dimensionality of the solution space is only $4M+6$. 

Next, define a set of surfaces ruled by the motion of lines in camera frame on a closed time interval. Let $X_{t_0}^{l}$, $V_{t_0}^{l}$ be the initial ruling position, direction, for the $l$-th ruling starting from time $t_0$, and $-{}^{t_0}X(t)$ be the camera translation signal originating at $t_0$ independent of individual lines. A set of ruled surfaces on interval $[t_0, t_n]$ is

\begin{equation} \label{eq:16}
RS(t_0, t_n) = \left\{ R(l, t_0, t) = X_{t0}^{l} + \alpha V_{t0}^{l} + {}^{t0}X(t) | t \in [t_0, t_n] \right\}_{l=1}^N
\end{equation}

Where $R(l, t_0, t)$ denotes the ruling of the $l$-th surface originating from time $t_0$ at time $t$, in which each line is optimized by Section \ref{sec:loss}. For example, Figure \ref{fig:1} illustrates the projected rulings in an image frame at some time. The surfaces ruled by these projections are shown in Figure \ref{fig:2}.

\begin{figure*}[!htb]
    \centering
    \begin{subfigure}[t]{0.32\textwidth}
        \centering
        \includegraphics[width=\textwidth]{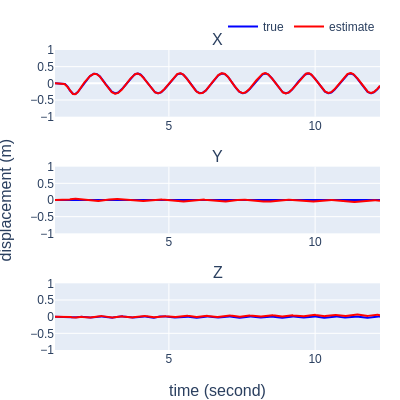}
        \caption{Trajectory}
        \label{fig:11_04_004_traj}
    \end{subfigure}
    \hfill
    \begin{subfigure}[t]{0.32\textwidth}
        \centering
        \includegraphics[width=\textwidth]{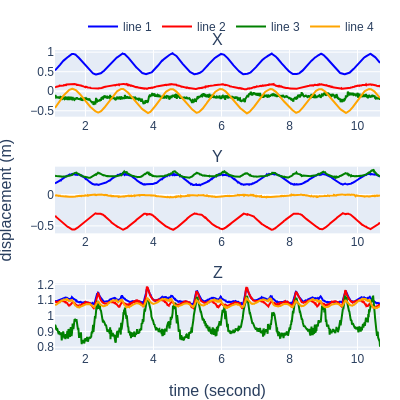}
        \caption{Directrices}
        \label{fig:11_04_004_x0}
    \end{subfigure}
    \hfill
    \begin{subfigure}[t]{0.32\textwidth}
        \centering
        \includegraphics[width=\textwidth]{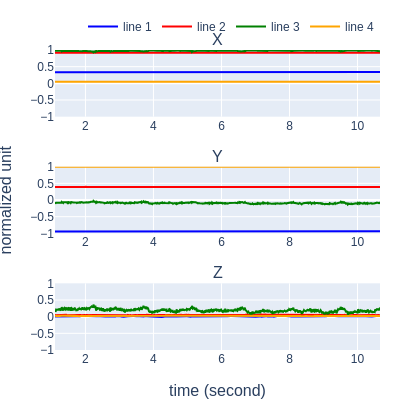}
         \caption{Rulings Directions}
        \label{fig:11_04_004_v0}
    \end{subfigure}
    
    \caption{(a) Estimated trajectory of the camera, (b) trajectory of the directrix for each ruled surface, and (c) ruling direction for each ruled surface, (a) in world coordinates with respect to the initial camera pose, (b)(c) in derotated camera coordinates; linear motion parallel to the scene plane.}
    \label{fig:11_04_004}
\end{figure*}

\section{Methodology} \label{method}

\subsection{Initial Surface Estimation}

The nature of the formulation makes the interest of investigation lies only on the straight edges. Inspired by event camera's exceptional capabilities at detecting edges in vision tasks \cite{Gallego2022}, the presented research establishes a procedure with image gradient to simulate ``events". 

Consider a 2D image that also translates across the time signal $t$, with the goal of identifying edges that form straight lines whose translation over $t$ will result in developable surfaces. By calculating the image gradient $\Delta I = \left[ \frac{\partial I}{\partial x}, \frac{\partial I}{\partial y}\right]$, which is obtained by applying convolution with Sobel filters, a thresholded edge map can be generated. From this edge map, straight lines are identified using the Hough transform.

Let $[t_0, t_0 + \delta t]$ represent a small time interval during which the change in image gradient along each line remains below a certain threshold $\epsilon$. The estimation of $N$ initial surfaces in equation \eqref{eq:10} is based on criteria derived from the quality of the Hough transform.
From there, the directions of the rulings in the projected image can also be calculated and used as part of the initial parameters.

\subsection{Extruding Surface into Future Frames}

For surfaces estimated on a time interval $[t_0, t_{i}]$, the algorithm has to extrude the same surface to a longer interval $[t_0, t_{i+1}]$. 

\begin{algorithm}
\caption{Extrude Surface}\label{alg:extrude}
\begin{algorithmic}
\State \textbf{Given:}
\State Estimated $RS(t_0, t_{i})$
\State Set $E$ of points projected by $RS(t_0, t_{i})$ on $[t_0, t_{i}]$
\State \textbf{Do:}
\State At $t_{i+1}$
\For{each line $l$}
\State Estimate ruling $R(l, t_0, t_{i+1})$ 
\For{each $p_i$ measured at $t_{i+1}$}
\If{$dist(p_i, R(l, t_0, t_{i+1})) < \tau$}
    \State Add $p_i$ to $E$
\EndIf
\EndFor
\EndFor
\While{Loss of $RS(t_0, t_{i+1})$ on $E > 0.01$}
\State Add small perturbation to $RS(t_0, t_{i+1})$ and run the optimizer again
\EndWhile
\State \textbf{Return:}
\State Estimated $RS(t_0, t_{i+1})$ from appended $E$
\end{algorithmic}
\end{algorithm}

As described in Algorithm \ref{alg:extrude}, if $(t_{i+1} - t_i) < \delta$ for some $\delta > 0$, such that the change in image gradient along each line remains below a threshold $\epsilon$, the surface at $t_{i+1}$ can be predicted with current estimation and acceleration measurements from the IMU. This prediction approximates a ruling $R(l, t_0, t_{i+1})$ based on the existing surface estimation $RS(t_0, t_i)$. 
Image points close to the estimated ruling within a certain distance, $dist(p_i, R(l, t_0, t_{i+1})) < \tau$, are sampled and appended to all points within that distance from the surface on $[t_0, t_i]$. This distance is the total least squares distance from sample point $p_i$ to the projection of the ruling $R(l, t_0, t_{i+1})$.

\begin{dmath}
    dist(p_i, R(l, t_0, t_{i+1})) = \frac{(r_2^y - r_1^y)p_i^x - (r_2^x - r_1^x)p_i^y + r_2^xr_1^y - r_2^yr_1^x}{\sqrt{(r_2^y - r_1^y)^2 + (r_2^x - r_1^x)^2}}\\
\end{dmath}

$where$

$r_1^x = \frac{[X_{t0}^{l}  + {}^{t0}X(t)]^x}{[X_{t0}^{l}  + {}^{t0}X(t)]^z}, r_1^y = \frac{[X_{t0}^{l} + {}^{t0}X(t)]^y}{[X_{t0}^{l}  + {}^{t0}X(t)]^z}$

$r_2^x = \frac{[X_{t0}^{l} + V_{t0}^{l} + {}^{t0}X(t)]^x}{[X_{t0}^{l} +  V_{t0}^{l} + {}^{t0}X(t)]^z}, r_2^y = \frac{[X_{t0}^{l} +  V_{t0}^{l} + {}^{t0}X(t)]^y}{[X_{t0}^{l} +  V_{t0}^{l} + {}^{t0}X(t)]^z}$

Here, a two-point formulation of distance calculation is used. Since $\alpha$ is free, the points $r_1, r_2$ can be easily calculated by setting $\alpha=0, 1$

The aggregated data is then used to estimate a new surface over the extended time interval, resulting in $RS(t_0, t_{i+1})$.
To improve this process, if the cost in the new estimation is greater than some small value (0.01), some random perturbation is added to the parameters until optimal solution with low cost is achieved. 
Repeating this process allows surface estimation to be extended into future frames.

\subsection{Sliding Window}

Due to the errors inherent to acceleration and gyro sampling on the hardware level, the estimation $RS(t_0, t_{n})$ would be less accurate as the length of the interval increases. This problem, however, can be mitigated by adopting a sliding window approach and gradually move the start time $t_0$ of the estimated surfaces. 

For a recording sampled for $\mathcal{T}$ seconds long at some frequency $\mathcal{F}$, there are $\mathcal{N} = \mathcal{T} \cdot \mathcal{F}$ images captured at unique timestamps. 
Let the start time of the $\mathcal{N}$-long sequence be $t_0$ and the end be $t_{\mathcal{N}}$, for any time interval $[t_j, t_{j+k}]$ where $j \geq 0$ and $(j+k) \leq \mathcal{N}$, define it to be a window $wind(k, t_j)$ of size $k$ starting at $t_j$ for some $k$ such that $(t_{j+k} - t_{j}) < \iota$ where $\iota$ is of reasonable length such that the hardware sampling error is negligible. 

\begin{algorithm}
\caption{Sliding Window}\label{alg:slide}
\begin{algorithmic}
\State \textbf{Given:}
\State Estimated $RS(t_j, t_{j+k})$ on $wind(k, t_j)$
\State Set $E$ of points projected by $RS(t_j, t_{j+k})$ on $wind(k, t_j)$
\State \textbf{Do:}
\State At $t_{j+k+1}$
\For{each line $l$}
\State Estimate ruling $R(l, t_j, t_{j+k+1})$ 
\For{each $p_i$ measured at $t_{j+k+1}$}
\If{$dist(p_i, R(l, t_j, t_{j+k+1})) < \tau$}
    \State Add $p_i$ to $E$
\EndIf
\EndFor
\EndFor
\State Trim $E$ to interval $[t_{j+1}, t_{j+k+1}]$
\While{Loss of $RS(t_{j+1}, t_{j+k+1})$ on $E > 0.01$}
\State Add small perturbation to $RS(t_{j+1}, t_{j+k+1})$ and run the optimizer again
\EndWhile
\State \textbf{Return:}
\State Estimated $RS(t_{j+1}, t_{j+k+1})$ from appended $E$
\end{algorithmic}
\end{algorithm}

As illustrated in Algorithm \ref{alg:slide}, starting with $j=0$ in window $wind(k, t_j)$ and estimated surfaces on that window $RS(t_j, t_{j+k})$, sliding the window and increasing $j$ repetitively, it can estimate the same surfaces on all windows $wind(k, t_j)$ for $j \geq 1$. 

\subsection{Recover Camera Odometry and Scene Geometry from Ruled Surface}

After collecting the surface estimations on each window from a sequence, the camera's trajectory, as well as surrounding scenes in terms of lines, can be reconstructed. While the latter is derived directly from the ruled surface as discussed in Section \ref{sec:4}, the former is less explicit. 

Let the camera translation signal $-{}^{t_i}X(t_j)$ denotes the displacement of camera at time $t_j$ relative to lines on its trajectory starting from $t_i$, the odometry ${}^{t_0}\hat{X}^{t_n}(t)$ from $t_0$ to $t_n$ is the summation of all consecutive displacements along the trajectory 

\begin{equation}
    -{}^{t_0}\hat{X}^{t_n}(t) = -\sum\limits_{i=0}^{n}{}^{t_i}X(t_{i+1})
\end{equation}

\begin{figure*}[!htb]
    \centering
    \begin{subfigure}[t]{0.32\textwidth}
        \centering
        \includegraphics[width=\textwidth]{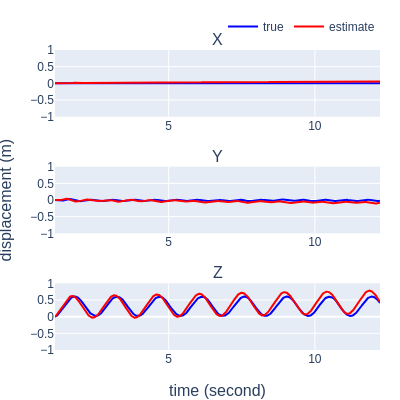}
        \caption{Trajectory}
        \label{fig:11_04_005_traj}
    \end{subfigure}
    \hfill
    \begin{subfigure}[t]{0.32\textwidth}
        \centering
        \includegraphics[width=\textwidth]{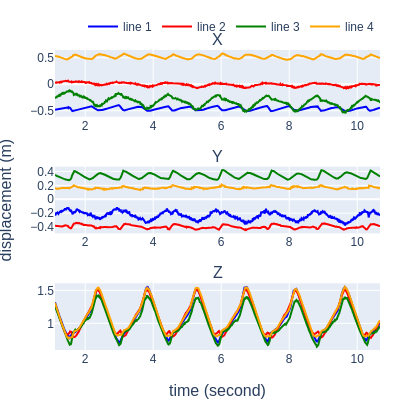}
        \caption{Directrices}
        \label{fig:11_04_005_x0}
    \end{subfigure}
    \hfill
    \begin{subfigure}[t]{0.32\textwidth}
        \centering
        \includegraphics[width=\textwidth]{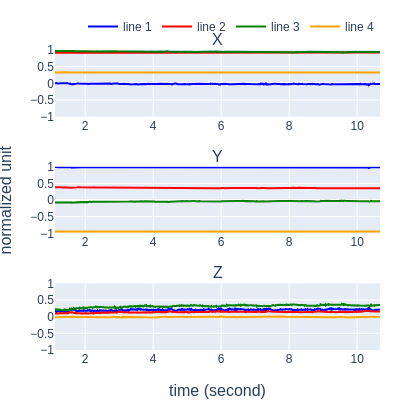}
        \caption{Ruling Directions}
        \label{fig:11_04_005_v0}
    \end{subfigure}
    
    \caption{(a) Estimated trajectory of the camera, (b) trajectory of the directrix for each ruled surface, and (c) ruling direction for each ruled surface, (a) in world coordinates with respect to the initial camera pose, (b)(c) in derotated camera coordinates; linear motion perpendicular to the scene plane.}
    \label{fig:11_04_005}
\end{figure*}

\begin{figure*}[!htb]
    \centering
    \begin{subfigure}[t]{0.32\textwidth}
        \centering
        \includegraphics[width=\textwidth]{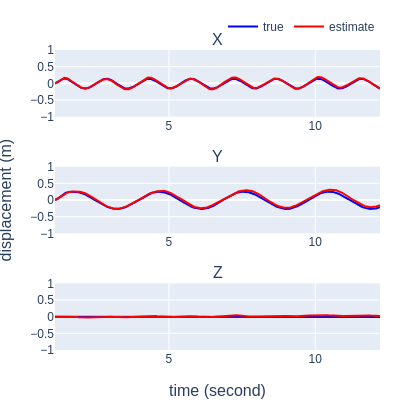}
        \caption{Trajectory}
        \label{fig:11_04_008_traj}
    \end{subfigure}
    \hfill
    \begin{subfigure}[t]{0.32\textwidth}
        \centering
        \includegraphics[width=\textwidth]{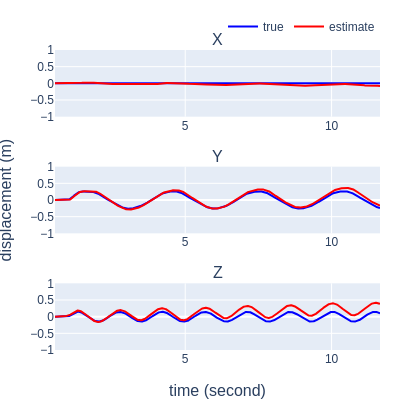}
        \caption{Trajectory}
        \label{fig:11_04_007_traj}
    \end{subfigure}
    \hfill
    \begin{subfigure}[t]{0.32\textwidth}
        \centering
        \includegraphics[width=\textwidth]{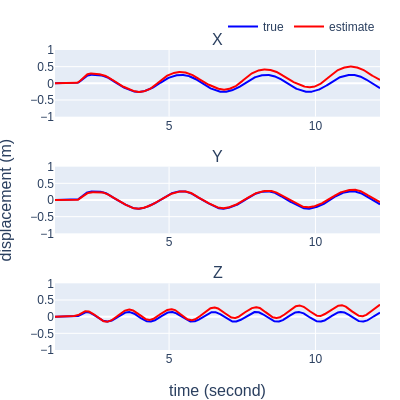}
        \caption{Trajectory}
        \label{fig:11_04_010_traj}
    \end{subfigure}
    
    \caption{Estimated camera trajectory under different circular motion configurations: (a) parallel to the scene plane, (b) perpendicular to the scene plane, and (c) tilted to the scene plane.}
    \label{fig:combined_circular_motion}
\end{figure*}

\section{Experiments}

\subsection{Setup}

Using a Realsense 435i camera running at $90$ fps with internal time-synchronized IMU, the algorithm is tested on recovering the correct camera trajectory on a series of scenes with different settings. The ground truth values of the trajectories are recovered in two ways: for the experiments with UR10 robot arm, the physical dimensions of the camera is collected from its design file such that the robot is set up in a way the trajectory of the tip of the arm is the same as the camera center up to millimeter accuracy; for the experiments with Vicon tracking, a joint calibration is done so that the rotation and translation from the Vicon world frame to the camera pose is always known, from which the ground truth camera odometry as well as the ground truth line poses are both known through Vicon.
Each sequence is roughly 12 seconds long.

\subsection{Surface Estimation}
To derive ruled surfaces, first chose a $0.1$ seconds long interval for initial line edges detection. The parameters for hough transform have $\rho = 5$ pixels and $\theta = 1$ degree with threshold for the accumulator being $100$, minimum line length being $100$ pixels and maximum line gap being $5$ pixels. Pick the line with the highest votes from the hough transform accumulator, remove the all data points that are within some distance (5 centimeters in undistorted image plane) to such line and run hough transform on the rest of the data points again. Repeat this process $M$ times result in $M$ initial surfaces ($N=3$ in the example shown in Figure \ref{fig:2}).

After the initial estimation, the surfaces can be extruded into the future with the sliding window technique mentioned above to eventually recover camera odometry and the environment's geometry. Based on observations setting window size to $140$ frames or roughly $1.55$ works well. The parameters are calculated through solving a nonlinear least-squares problem that minimizes the ruling reprojection loss of equation \eqref{eq:12}.

\subsection{Derotation}
To achieve the requirement of rotation-less camera, it is critical to derotate the camera frames with gyro measurements from the IMU sensor, which can be integrated into quaternion units.

Let rotation matrix ${}^{t_2}R^{t_1}$ denotes the rotation of the camera pose from time $t_1$ to $t_2$, for mitigating the errors introduced in gyro integration, it needs to match the same sliding window approach where the frames within a single window are de-rotated to the same frame at $w_0$. Denote this $w_0$ as the rotation center of the window.

For a $k$-length window, for each frame at $w_i, 0<i \leq k$, rotate it by ${}^{w_0}R^{w_i}$ to negate the rotation of camera within the window. However, since this process makes the camera and line poses in each window unique to their rotation, another procedure to rotate each window to some common rotation center $t_0$ is naturally needed. For a window $wind(k, w_j)$ starting at some time $w_j$, apply rotation ${}^{t_0}R^{w_j}$ to frames and estimated poses within the window to combine the windows. To restore the rotation, a final rotation ${}^{t_i}R^{t_0}$ can be applied to each frame where $t_i$ is their respective time stamp.

\begin{figure*}[!htb]
    \centering
    \begin{subfigure}[t]{0.32\textwidth}
        \centering
        \includegraphics[width=\textwidth]{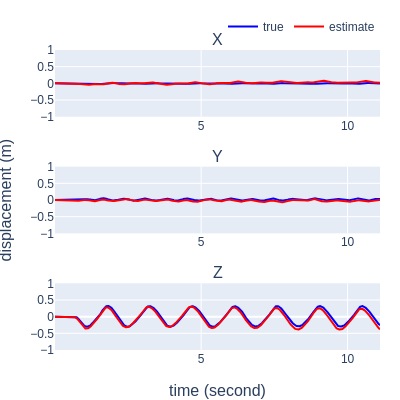}
        \caption{Trajectory}
        \label{fig:11_01_003_traj}
    \end{subfigure}
    \hfill
    \begin{subfigure}[t]{0.32\textwidth}
        \centering
        \includegraphics[width=\textwidth]{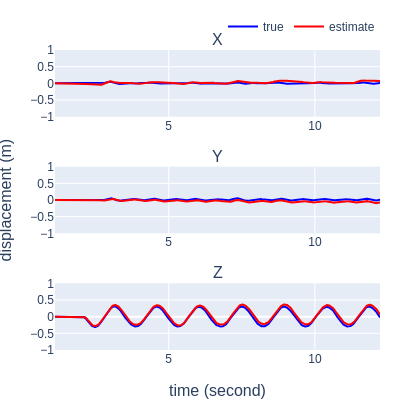}
        \caption{Trajectory}
        \label{fig:11_02_007_traj}
    \end{subfigure}
    \hfill
    \begin{subfigure}[t]{0.32\textwidth}
        \centering
        \includegraphics[width=\textwidth]{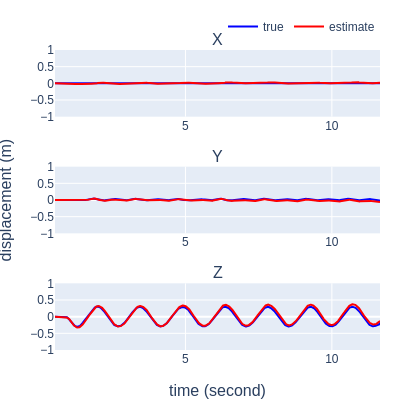}
        \caption{Trajectory}
        \label{fig:11_01_005_traj}
    \end{subfigure}
    
    \caption{Estimated camera trajectories with linear motion perpendicular to the scene plane, combined with rotation in (a) the X direction, (b) the Y direction, and (c) the Z direction of the camera.}
    \label{fig:combined_linear_rotation}
\end{figure*}

\begin{figure*}[!htb]
    \centering
    \begin{subfigure}[t]{0.32\textwidth}
        \centering
        \includegraphics[width=\textwidth]{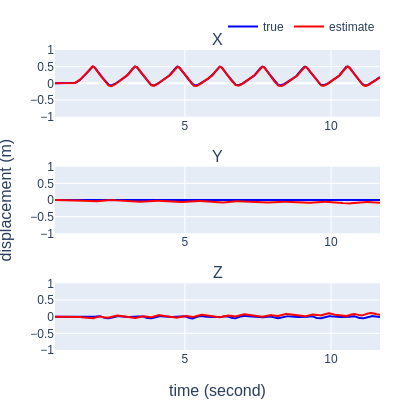}
        \caption{Trajectory}
        \label{fig:11_04_022_traj}
    \end{subfigure}
    \hfill
    \begin{subfigure}[t]{0.32\textwidth}
        \centering
        \includegraphics[width=\textwidth]{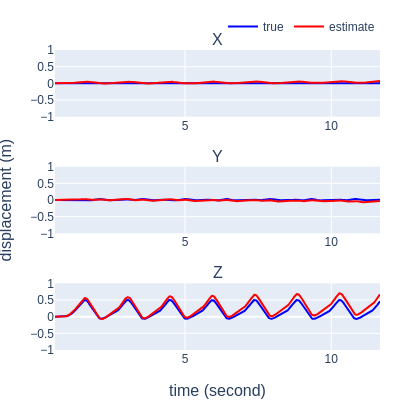}
        \caption{Trajectory}
        \label{fig:11_04_025_traj}
    \end{subfigure}
    \hfill
    \begin{subfigure}[t]{0.32\textwidth}
        \centering
        \includegraphics[width=\textwidth]{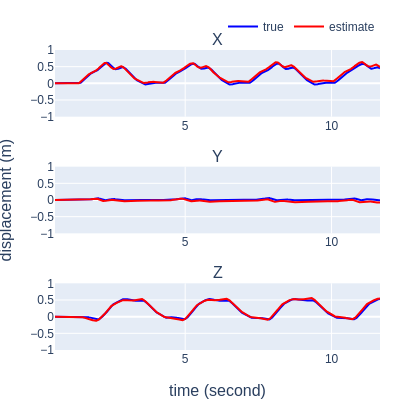}
        \caption{Trajectory}
        \label{fig:11_05_000_traj}
    \end{subfigure}
    
    \caption{Estimated camera trajectories with linear non-smooth motion in different orientations relative to the scene plane: (a) parallel, (b) perpendicular, and (c) square.}
    \label{fig:combined_non_smooth_motion}
\end{figure*}

\begin{figure*}[!htb]
    \centering
    \begin{subfigure}[t]{0.32\textwidth}
        \centering
        \includegraphics[width=\textwidth]{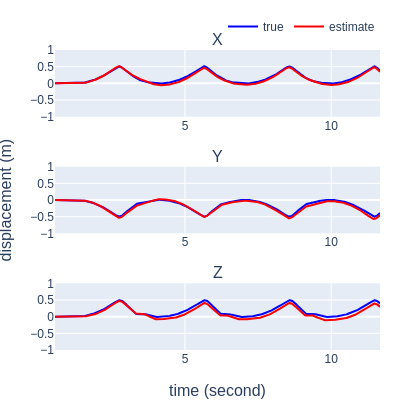}
        \caption{Trajectory}
        \label{fig:11_04_027_traj}
    \end{subfigure}
    \hfill
    \begin{subfigure}[t]{0.32\textwidth}
        \centering
        \includegraphics[width=\textwidth]{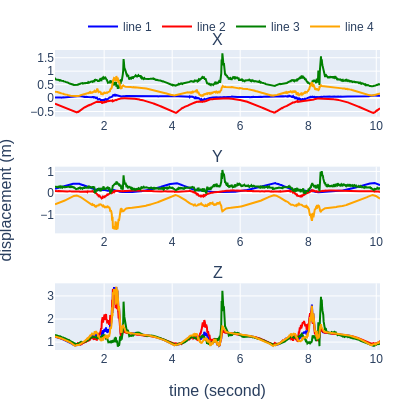}
        \caption{Directrices}
        \label{fig:11_04_027_x0}
    \end{subfigure}
    \hfill
    \begin{subfigure}[t]{0.32\textwidth}
        \centering
        \includegraphics[width=\textwidth]{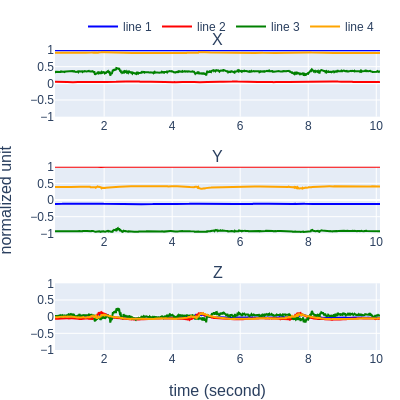}
        \caption{Ruling Directions}
        \label{fig:11_04_027_v0}
    \end{subfigure}
    
    \caption{(a) Estimated trajectory of the camera, (b) trajectory of the directrix for each ruled surface, and (c) ruling direction for each ruled surface, (a) in world coordinates with respect to the initial camera pose, (b)(c) in derotated camera coordinates; non-smooth linear motion tilted to the scene plane.}
    \label{fig:11_04_027}
\end{figure*}

\section{Results}

Starting from simple motions, first consider coplanar lines (lines formed in a square shape as in Figure \ref{fig:10_31_028_1}) and single directional rotation-less motions, parallel or perpendicular to the plane of the lines. 
On top of linear motions, circular ones with sinusoidal signals in two or more directions are also processed, illustrated in Figure \ref{fig:11_04_008_4}. 

Naturally, any real world camera motion would involve rotations. The motions with rotations are grouped into two sets, linear and circular, again similar to the previous setup. To test the robustness of the algorithm, for each motion rotations in all three directions, X, Y, Z, relative to camera itself, are added separately. 
If the translation is in one direction, this setup is intended to test if the rotation in any direction would disproportionately affect the evaluation quality. 

Other than only smooth, sinusoidal motions are tested on, it is sensible to study the cases there the motion is not smooth as well. 
In the same form of back-and-forth motions like Figure \ref{fig:10_31_028_4}, instead of sinusoidal change of positions, a zigzag signal is used. And on top of those movements, motion in the shape of a square, like Figure \ref{fig:11_05_006_4}, with abrupt stopping at each corner is also used.

Advancing from simple coplanar lines, tests are performed on more complicated non-coplanar scenes to test if the algorithm generalizes. 
Figure \ref{fig:11_06_008_all} sums up the setup for this experiment. The four lines being tracked are carefully placed to be non-coplanar which generates a more convoluted scene when estimating \eqref{eq:5}. Since the lines are not coplanar, all motions are tilted in this context. 

Finally, to test beyond idealized motions, tests were done with hand held camera where ground truth odometry is recorded with Vicon tracking system. 
The scene is setup as illustrated in Figure \ref{fig:1} and both coplanar and non-coplanar scenes are tested. 
The camera is hand-held by the tester who moves it freely in front of the scene. Markers for tracking the poses of objects are attached to the lines in the scene too so that the Vicon system can record the ground truth poses for lines in world coordinates.
Running a joint calibration between the Vicon system and the camera, a transformation from Vicon world frame to camera frame can be obtained, from which the distance between markers and the estimated lines in camera coordinates is calculated in ensuing experiments.

\begin{table}[!htb]
    \centering
    \resizebox{\textwidth/2}{!}{%
    \begin{tabular}{@{}lccc@{}}
    \toprule
    \textbf{Motion} & \textbf{X (m)} & \textbf{Y (m)} & \textbf{Z (m)} \\
    \midrule
    \multirow{1}{*}{Linear Parallel} & 0.0019	± 0.0080 &  0.0138 ±	0.0215 & 0.0251 ± 0.0173\\
    \multirow{1}{*}{Linear Perpendicular} & 0.0251 ± 0.0159 & 0.0341 ± 0.0270 & 0.0554 ± 0.1004 \\
    \midrule
    \multirow{1}{*}{Circular Parallel}  
     & 0.0097 ± 0.0197 & 0.0320 ± 0.0230 & 0.0159 ± 0.0211 \\
    \multirow{1}{*}{Circular Perpendicular}  
     & 0.0228 ± 0.0258 & 0.0317 ± 0.0384 & 0.1201 ± 0.0800 \\
    \multirow{1}{*}{Circular Tilted-Axis} & 0.1038 ± 0.0808 & 0.0163 ± 0.0236 & 0.1049 ± 0.0665 \\
    \bottomrule
    \end{tabular}%
    }
    \caption{Difference between True and Estimated Trajectories, Rotation-less Motion}
    \label{tab:1}
\end{table}

\begin{table*}[!htb]
    \centering
    \resizebox{0.8\textwidth}{!}{%
    \begin{tabular}{@{}lcccc@{}}
    \toprule
    \textbf{Motion} & \textbf{Rotation Direction}& \textbf{X (m)} & \textbf{Y (m)} & \textbf{Z (m)} \\
    \midrule
     \multirow{3}{*}{Linear Parallel} & X & 0.0278 ± 0.0289 & 0.0171 ± 0.0369 & 0.0041 ± 0.0086 \\
     & Y & 0.0510 ± 0.0446 & 0.0223 ± 0.0170 & 0.0874 ± 0.0607 \\
    & Z & 0.0035 ± 0.0115 & 0.0452 ± 0.0240 & 0.0222 ± 0.0104 \\
    \midrule
    \multirow{3}{*}{Linear Perpendicular} & X & 0.0130 ± 0.0260 & 0.0310 ± 0.0144 & 0.0461 ± 0.0421 \\
    & Y & 0.0211 ± 0.0305 & 0.0395 ± 0.0230 & 0.0427 ± 0.0268 \\
    & Z & 0.0064 ± 0.0117 & 0.0164 ± 0.0130 & 0.0313 ± 0.0306 \\
    \midrule
    \multirow{3}{*}{Circular Parallel} & X &0.0163 ± 0.0156 &  0.0438	± 0.0252 & 0.0229 ± 0.0189\\
    & Y & 0.0052 ± 0.0204 & 0.0285 ± 0.0247 &  	0.0381 ± 0.0255\\
    & Z & 0.0005 ± 0.0254	& 0.0195 ± 0.0182 &	0.0270 ± 0.0164 \\
    \midrule
    \multirow{3}{*}{Circular Perpendicular} & X & 0.0012 ± 0.0174 & 0.0222 ± 0.0249 & 0.0002 ± 0.0141 \\
    & Y & 0.0207 ± 0.0250 & 0.0510 ± 0.0286 & 0.0098 ± 0.0211 \\
    & Z & 0.0528 ± 0.0286 & 0.0283 ± 0.0187 & 0.0616 ± 0.0431 \\
    \midrule
    \multirow{3}{*}{Circular Tilted} & X & 0.0068 ± 0.0202 & 0.0933 ± 0.0552 & 0.1125 ± 0.0846 \\
    & Y & 0.0836 ± 0.0672 & 0.0039 ± 0.0102 & 0.0272 ± 0.0182 \\
    & Z & 0.0856 ± 0.0510 & 0.0103 ± 0.0136 & 0.0957 ± 0.0439 \\
    \bottomrule
    \end{tabular}%
    }
    \caption{Difference between true and estimated trajectories, motion with rotation. }
    \label{tab:2}
\end{table*}

\section{Discussion}

\subsection{Rotation-less Motions}\label{sec:r1}

Figure \ref{fig:10_31_028_4}, for example, is a trajectory of a slice of roughly 12 seconds in world frame, its equivalent trajectory relative to the camera initial pose on X,Y,Z directions is plotted in Figure \ref{fig:11_04_004_traj} while Figure \ref{fig:11_04_005_traj} shows the perpendicular case. 
While the trajectory is accurately recovered, a drift caused by cumulating IMU measurement errors is visible particularly in the latter graph. 
This is also evident in Table \ref{tab:1} where the perpendicular motion has higher mean and standard deviation on difference from ground truth.

The directrix (Figures \ref{fig:11_04_004_x0}, \ref{fig:11_04_005_x0}) and ruling (Figures \ref{fig:11_04_004_v0}, \ref{fig:11_04_005_v0}) in derotated camera coordinates can also be recovered. 
Since the lines are organized in a roughly square formation, it is reflected in parallel motion (Figure \ref{fig:11_04_004_x0}) where two of the lines move similarly to the camera motion while the other two don't as the two sides of the square that are parallel to the motion is not moving in the sense of line orientation. 
On the other hand, for the perpendicular motion (Figure \ref{fig:11_04_005_x0}), the sinusoidal motion in all lines can be observed with the depth of the lines changing in uniform. Note that the trajectory of the directrix which portrays partly the motion of the line in camera coordinates is not just influenced by the camera motion, it is also forced such that the vector from camera origin to the directrix is perpendicular to the ruling for dimensionality reduction purposes as discussed in Section \ref{sec:4_2}.
The ground truth of the line poses, which are compared in a later section, are not available in this set of experiments. 

Since the motion is supposedly rotation-less, the ruling should remain relative constant across time; as revealed in Figures \ref{fig:11_04_004_v0}, \ref{fig:11_04_005_v0} the estimated results have different levels of noises, particularly troublesome in the Z direction.  
Albeit the camera trajectory is well constructed, the line scene geometry is less stable. 
The X,Y directions of the rulings are initialized vie the projected direction of the lines in the image while the Z direction is unknown. This is the reason why the algorithm produces more noise on estimating ruling's Z direction.  
When comparing the estimated trajectories to the ground truth, as shown in Figure \ref{fig:combined_circular_motion}, it is clear that the perpendicular motions are worse than the parallel one.
This is most likely caused by the drift of acceleration measurements on the IMU censor, possibly in Z direction more than the other directions.

\subsection{Motions with Rotations}

Again, motion in linear and circular forms are tested with motions parallel, perpendicular, and tilted with respect to the lines plane.
The quantitative results are listed in Table \ref{tab:2}. 

Rotation in any direction does not seem to impose any negative effects on the estimated odometry, it seems to perform better than the rotation-less cases as the drift of acceleration is dispersed rather than concentrated in one direction. 
For example, in Figure \ref{fig:11_04_005_traj} it was tested that there's a drift in the camera Z direction that has been negatively impacting the estimation. If the rotation is added, however, the algorithm was able to predict more accurate results as shown in Figure \ref{fig:combined_linear_rotation}. This is because while the linear acceleration of the trajectory in world frame is roughly the same, the rotation is applied to the acceleration measured by the IMU.
Since the error is only persistent in the Z direction value reported by the IMU, this error is dispersed to other directions when the acceleration measurements are derotated to some basis frame, on which the trajectory of camera is based.
Thus, the errors accumulated in the Z direction of the actual trajectory is reduced.  
Similar improvements can also be seen in other types of motion. 

\begin{figure*}[!htb]
    \centering
    \begin{subfigure}[t]{0.32\textwidth}
        \centering
        \includegraphics[width=\textwidth]{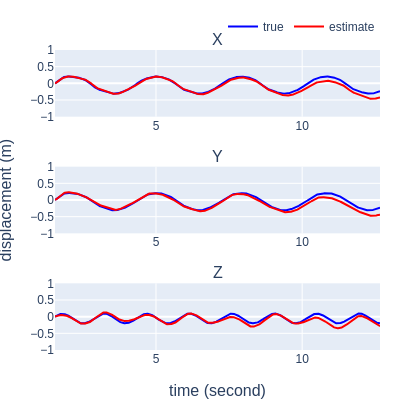}
        \caption{Trajectory}
        \label{fig:11_06_008_traj}
    \end{subfigure}
    \hfill
    \begin{subfigure}[t]{0.32\textwidth}
        \centering
        \includegraphics[width=\textwidth]{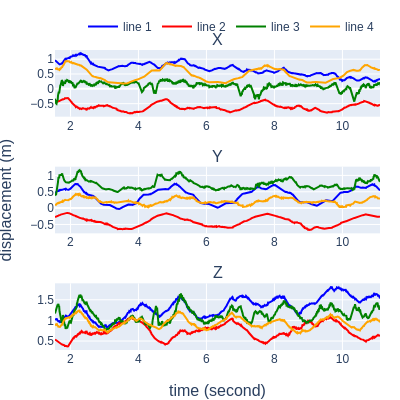}
        \caption{Directrices}
        \label{fig:11_06_008_x0}
    \end{subfigure}
    \hfill
    \begin{subfigure}[t]{0.32\textwidth}
        \centering
        \includegraphics[width=\textwidth]{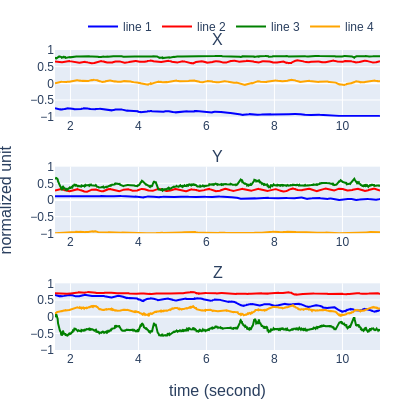}
        \caption{Ruling Directions}
        \label{fig:11_06_008_v0}
    \end{subfigure}
    
    \caption{(a) Estimated trajectory of the camera, (b) trajectory of the directrix for each ruled surface, and (c) ruling direction for each ruled surface, (a) in world coordinates with respect to the initial camera pose, (b)(c) in derotated camera coordinates with rotating circular motion on non-coplanar scenes.}
    \label{fig:11_06_008}
\end{figure*}

\begin{table}[!htb]
    \centering
    \resizebox{\textwidth/2}{!}{%
    \begin{tabular}{@{}lccc@{}}
    \toprule
    \textbf{Non-smooth Motion} & \textbf{X (m)} & \textbf{Y (m)} & \textbf{Z (m)} \\
    \midrule
    \multirow{1}{*}{Linear Parallel} & 
    0.0081 ± 0.0071 &  0.0504 ± 0.0267 & 0.0317 ± 0.0319\\
    \multirow{1}{*}{Linear Perpendicular} & 0.0241 ± 0.0206 & 0.0204 ± 0.0202 & 0.0957 ± 0.0558 \\
    \multirow{1}{*}{Linear Tilted}  & 
     0.0255 ± 0.0221 & 0.0259 ± 0.0266 & 0.0631 ± 0.0362 \\
     \midrule
    \multirow{1}{*}{Square Parallel} & 
    0.0850 ± 0.0478 &  0.0138 ± 0.0182 & 0.0626 ± 0.0427\\
    \multirow{1}{*}{Square Perpendicular} & 0.0096 ± 0.0217 & 0.0126 ± 0.0077 & 0.0449 ± 0.0549 \\
    \bottomrule
    \end{tabular}%
    }
    \caption{Difference between true and estimated trajectories, non-smooth motion.}
    \label{tab:3}
\end{table}

\begin{table}[!htb]
    \centering
    \resizebox{\textwidth/2}{!}{%
    \begin{tabular}{@{}lccc@{}}
    \toprule
    \textbf{Motion} & \textbf{X (m)} & \textbf{Y (m)} & \textbf{Z (m)} \\
    \midrule
    \multirow{1}{*}{Circular} & 0.0347 ± 0.0404 & 0.0189 ± 0.0387 & 0.2237 ± 0.1428 \\
    \multirow{1}{*}{Circular w/ Rotation}  & 
     0.0410 ± 0.0531 & 0.0423 ± 0.0578 & 0.0373 ± 0.0543 \\
    \bottomrule
    \end{tabular}%
    }
    \caption{Difference between true and estimated trajectories, non-coplanar scene.}
    \label{tab:4}
\end{table}

\begin{table*}[!htb]
    \centering
    \resizebox{0.8\textwidth}{!}{%
    \begin{tabular}{@{}lccccc@{}}
    \toprule
    \textbf{Scene} & \textbf{X (m)} & \textbf{Y (m)} & \textbf{Z (m)} & \textbf{Line 1 (m)} & \textbf{Line 2 (m)}\\
    \midrule
    \multirow{1}{*}{Coplanar 1} & 0.0236 ± 0.0147 & 0.0257 ± 0.0084 & 0.0207 ± 0.0140 & 0.0597 ± 0.0290 & 0.0591 ± 0.0339\\
    \multirow{1}{*}{Coplanar 2}  & 
     0.0139 ± 0.0492 & 0.0519 ± 0.0785 & 0.0600 ± 0.0749 & 0.2111 ± 0.1940 & 0.2061 ± 0.1412\\
     \multirow{1}{*}{Non-Coplanar 1} & 0.0821 ± 0.0642 & 0.0181 ± 0.0919 & 0.1968 ± 0.1301 & 0.3713 ± 0.2017 & 0.3885 ± 0.2390\\
    \multirow{1}{*}{Non-Coplanar 2}  & 
     0.2441 ± 0.1387 & 0.0169 ± 0.0182 & 0.1739 ± 0.0614 & 0.3935 ± 0.1586 & 0.3419 ± 0.1990\\
    \bottomrule
    \end{tabular}%
    }
    \caption{Difference between true and estimated trajectories and line poses, under free motion.}
    \label{tab:5}
\end{table*}

\subsection{Non-smooth Motions} \label{sec:r3}
The estimated and true trajectories are shown in Figure \ref{fig:combined_non_smooth_motion}. Notice that the same drift still persist on camera Z axis. Moreover, when considering the tilted motion shown in Figure \ref{fig:11_04_027}, the estimated directrices have more drastic movements as in Figure \ref{fig:11_04_027_x0} and the rulings have more noise as in Figure \ref{fig:11_04_027_v0}. 
This, however, does not equate to bad visual odometry estimation in Figure \ref{fig:11_04_027_traj}, and does not necessarily equate to bad scene estimation either, as the directrix can move along the axis of the line itself. 
Analyzing the errors quantitatively, in Table \ref{tab:3}, there is not a negative impact of the non-smoothness on the sequences. It can be seen that the algorithm fares well on even the sporadic motions.

\subsection{Non-coplaner Scenes}\label{sec:r4}

Investigating the errors in Table \ref{tab:4}, it can be said that the algorithm achieves comparably accuracy on the non-coplanar scenes. 
The drift on the Z axis is once again observed and adding rotations to the motion improves the estimated odometry.  
Each directrix has different motion, manifested in Figure \ref{fig:11_06_008_x0}, as the rulings are non-coplanar. Examining Figure \ref{fig:11_06_008_v0}, line 1 is drifting disproportionately in the Z direction comparing the other lines. 
This does signify a potential discrepancy between the estimated and true line pose, though the latter of which is not known in this part of the experiments.
Nevertheless, the impact of one erroneous ruled surface can be shown to have limited impact on the odometry estimate.
Taking a look at Figure \ref{fig:11_06_008_traj}, one can get a sense of how well the camera's trajectory was reconstructed.

\begin{figure*}[!htb]
    \centering
    \begin{subfigure}[t]{0.45\textwidth}
        \centering
        \includegraphics[width=\textwidth]{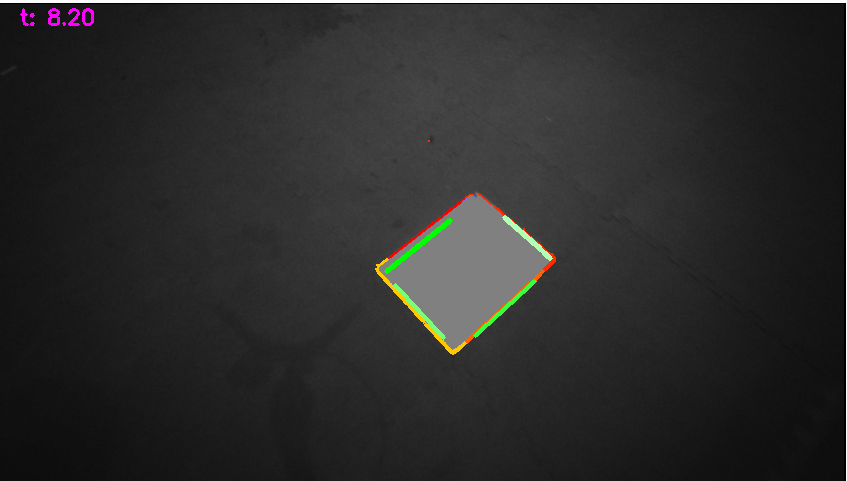}
        \caption{Coplanar Scene, Clean Background}
        \label{fig:09_23_003_1}
    \end{subfigure}
    \hfill
    \hfill
    \begin{subfigure}[t]{0.45\textwidth}
        \centering
        \includegraphics[width=\textwidth]{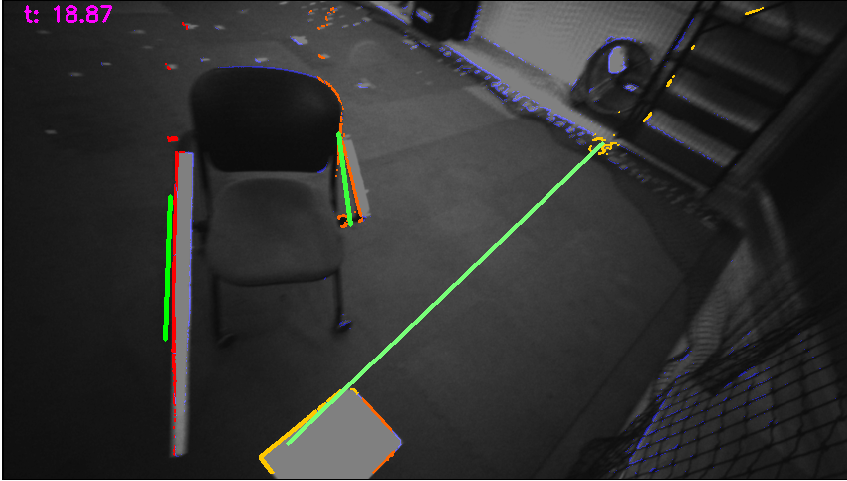}
        \caption{Non-Coplanar Scene, Noisy Background}
        \label{fig:09_23_002_1}
    \end{subfigure}
    
    \caption{(a) Coplanar scene and (b) non-coplanar scene used in Section \ref{sec:r5}. The algorithm performs well on the former but struggles on the latter. Since the rulings are considered unbounded noisy points in the background negatively impacts the estimation.}
    \label{fig:limitation}
\end{figure*}

\subsection{Free Motions}\label{sec:r5}

Moving on from the previous setup, in order to see how well the presented research works in less idealized environment, the investigations in this section are done to test sequences with free movements with ground truth from Vicon motion capture system. The estimation errors are listed in Table \ref{tab:5}. 
While the algorithm achieved similar accuracy as previous experiments in the coplanar scene the estimation on the non-coplanar ones are the worst so far. 
It can be said that it is more difficult to fit surfaces ruled by non-coplanar lines with irregular motions but considering neither non-smooth motions in Section \ref{sec:r3} nor non-coplanar scenes in Section \ref{sec:r4} had any significant impact on the estimation accuracy, it is not known what exactly is hindering the performance. 

As for the accuracy of the estimated line poses, the distance from Vicon markers that are supposedly on the estimated lines to those lines in 3D are computed. 
Each line is typically marked by 2 to 4 markers and the average between them are taken as the reported result. 
For each scene two lines are tracked by the Vicon system. 
In general the quality of line poses is lower comparing to the relatively accurate camera motion, shown in Table \ref{tab:5}. 
In the scene of Coplanar 2, it seems even poor (20cm difference) line poses have only limited effects on the trajectory estimate (3cm increase in error).
On the other hand, poor estimates in non-coplanar scenes are accompanied by erroneous line poses estimation. 

\subsection{Limitations}

Given all the experiments conducted, the limitations of the presented research is discussed. 
As evident in Section \ref{sec:r5}, the algorithm is far from perfection when tested on free motions with non-coplanar scenes, which are more akin to scenarios in the real world. 
Here it is argued the decrease in accuracy is not a result of non-coplanarity or non-smoothness of motions, but rather happens during surface growing phase. 
Demonstrated in Figure \ref{fig:limitation}, as the current approach views the rulings as unbounded lines, noisy scene and outliers points belonging to other lines can easily interfere the sliding windows when new frames are added, which in turn further decreases the estimation accuracy in the next fitting step at the end of Algorithm \ref{alg:slide}.  
Resolving this issue, however, is not as straight forward as one might think since allowing the line to be unbounded reduces its degrees of freedom as discussed in Section \ref{sec:4_2} and attempts to associate points with lines may introduce correspondences. 

For future work, it is interesting to extend the algorithm to curves not just straight lines. Using event camera for better edge detection is also a possible branch of ways to improve overall performance.

\section{Conclusion}
The presented research designed a novel algorithm to estimate visual odometry from ruled surfaces. Instead of relying on point-to-point or line-to-line correspondence like traditional approaches, the work exploits the geometric properties of ruled surfaces to differentially update them.
To reduce the dimensionality of the solution space, measurements from a IMU sensor are used, constraining the estimated motion. 
Various experiments of different types of motions and scenes are conducted to evaluate the accuracy of the algorithm empirically.


\section*{Acknowledgments}
The support of Maryland Robotics Center and the NSF under award OISE 2020624 is gratefully acknowledged.



\end{document}